\documentclass{article}

\usepackage{times}
\usepackage{epsfig}
\usepackage{graphicx}
\usepackage{amsmath}
\usepackage{amssymb}

\usepackage[title]{appendix}

\usepackage[linesnumbered,ruled,vlined]{algorithm2e}
\SetArgSty{textnormal}
\usepackage{caption} 
\usepackage{amsfonts}
\usepackage{subfigure}
\usepackage{longtable}
\usepackage{multirow}
\usepackage{booktabs}
\usepackage[pagebackref=true,breaklinks=true,colorlinks,bookmarks=false]{hyperref}

\begin{document}
\newcommand\kai[1]{{\color{black}#1}}
\newcommand\dy[1]{{\color{black}#1}}
\newcommand\qh[1]{{\color{black}#1}}
\newcommand\zs[1]{{\color{black}#1}}
\title{Toward Better Generalization in Few-Shot Learning through \\the Meta-Component Combination\footnote{This work was conducted five years ago. Since then, we have observed that many researchers have arrived at similar ideas, which motivates us to share it here. We also believe these ideas are now amenable to integration into large language models.}}

\author{Qiuhao Zeng\\
}
\date{}
\maketitle
 
\begin{abstract}

In few-shot learning, classifiers are expected to generalize to unseen classes given only a small number of instances of each new class. One of the popular solutions to few-shot learning is metric-based meta-learning. However, it highly depends on the deep metric learned on seen classes, which may overfit to seen classes and fail to generalize well on unseen classes. To improve the generalization, we explore the substructures of classifiers and propose a novel meta-learning algorithm to learn each classifier as a combination of meta-components. Meta-components are learned across meta-learning episodes on seen classes and disentangled by imposing an orthogonal regularizer to promote its diversity and capture various shared substructures among different classifiers. Extensive experiments on few-shot benchmark tasks show superior performances of the proposed method.

\end{abstract}

\section{Introduction}
\label{introduction}



Recently, the success of deep learning has been witnessed in vision~\cite{krizhevsky2012imagenet}, language~\cite{vaswani2017attention} and speech~\cite{graves2013speech} areas. Though prevailing, deep learning algorithms perform unsatisfactorily on data with limited labels. To resolve this issue, few-shot learning~\cite{fei2006one,lake2011one,miller2000learning} is proposed to recognize unseen categories from very few labeled instances. The recent focus of few-shot learning is built on the meta-learning framework~\cite{finn2017model, munkhdalai2017meta,ravi2016optimization}. 

Meta-learning learns to quickly adapt to a novel task based on a few instances by leveraging the knowledge from previous related tasks. One group of the prevailing meta-learners is the metric-based approach~\cite{koch2015siamese,snell2017prototypical,sung2018learning,vinyals2016matching}, which learns a similarity metric shared by all instances in all classes. Another group is the optimization-based approach\cite{finn2017model, nichol2018first, ravi2016optimization}. Optimization-based methods adjust the optimization algorithm such that the model can be fast adapted with a few instances. 

In most metric-based approaches, the prediction is made by using the nearest neighbor classifiers. As the nearest neighbor classifier is learning-free, the performance highly depends on the learned metric. Generally, the metric learned in meta-learning may overfit to class-level structures of seen classes and fail to generalize well on unseen classes in a novel task. Suppose we have a metric that is trained on a fruit-classification task. The trained metric will assign extremely small distances between the images of ``red apple'' and ``green apple'' as they belong to the same class ``apple''  and large distances between images of ``red apple'' and ``red strawberry'' as they are in different classes. Suppose there comes a novel task which is to classify the color of fruits. In this task, using metric learned from the previous task can hardly separate the images of ``red apple'' and ``green apple'' but mistakenly separate the ``red apple'' and ``red strawberry''.


%
The major cause of the generalization failure of the metric learned from previous tasks is that classifiers fail to capture the change of subclass-level structures of each class in different tasks. In the above example, the images of ``apples'' may have three subclass-level structures: the structure of the fruit ``apple'' and two structures of color ``red'' and ``green''. 
When the task is to classify fruit categories, the classifier should highlight the structure of fruits and pay less attention to the structures of colors.~\footnote{Note that we cannot completely ignore the structures of colors as it contributes to distinguishing an ``apple'' from other fruits, such as a ``blueberry'', whose color is different from apples'.} When the task is to classify color categories, the classifier should focus more on the structures of colors and pay less attention to the structures of fruits. However, the metric trained in the class level cannot adapt its attention to task-relevant subclass-level structures with an extremely small set of training instances.

To address this issue, we propose to generate a classifier or a predictive model as a weighted combination of different components. Each component is responsible to classify a subclass-level structure and the components are shared along with all tasks. The combination scores are generated based on the support instances in a task and are expected to capture the changeable importance of the subclass-level structures in different tasks.
Note that we do not assume any prior knowledge about the subclass-level structure or the components. Instead, we learn the components by the meta episode training with an orthogonality-promoting regularizer~\cite{brock2018large,xie2018orthogonality} to encourage the diversity of the components, capturing both the frequent and infrequent subclass-level structures.


Another main cause of the generalization failure for metric-based approaches is the absence of adaptations to given instances in unseen classes. As the metric is fixed for novel tasks, it cannot update its parameters to explore the novel information of unseen classes. To better fit the task-specific information, we take the advantage of the optimization-based meta-learner via adapting (fine-tuning) the combination scores of the components in each classifier. Hopefully, in the example of the task to classify the fruit categories, the scores would be adapted to highlight the structures of the fruit and downplay the structures of colors. 

The contributions of the paper are summarized as follows:
\begin{itemize}
  \item[1)] A novel meta-learning approach, named Meta Components Learning (MCL), is proposed to tackle the classification generalization problem on unseen classes. MCL meta-learns the components of the classifiers. Then, the constructed classifiers are combined with all weighted components based on the support instances. Specially, we apply an orthogonality-promoting regularizer to facilitate the learning of components to capture more information in the classifiers.
  
  \item[2)] The generalization ability of MCL in novel tasks is improved by proposing to adapt the combination scores of the components for each predictive model, which helps the classifiers capture the class information in the support instances.
  
  \item[3)] The generalization of our method is tested on standard few-shot learning benchmarks against state-of-the-art methods. The results verify the superior performances of the proposed method for supervised learning and reinforcement learning tasks. 
\end{itemize}

\section{Related Work}


Few-shot learning aims to recognize novel classes with insufficient labeled instances. The recent focus of few-shot learning algorithms is the meta-learning framework~\cite{finn2017model}, which trains the model with multiple few-shot tasks. The meta-learning methods related to our approach fall into two categories: metric learning-based methods and optimization-based methods.\\

\noindent \textbf{Metric learning-based methods.} In these methods, a metric is learned to compare two instances' feature embeddings encoded by the feature extractor. Then an unseen image is classified based on its distances to the labeled instances in the supporting set. Prototypical Networks~\cite{snell2017prototypical} performs classification by computing distances to prototype representations of each class. SNAIL~\cite{ye2020few} uses soft attention to aggregate information from past experiences. FEAT~\cite{ye2020few} customizes a task-specific embedding space via a self-attention architecture. 


Our method is closer to LwoF~\cite{gidaris2018dynamic} and COMET~\cite{cao2020concept}. LwoF~\cite{gidaris2018dynamic} generates the classifier for new class by attention-based weighted combination on the classifiers of seen classes, also called base classes. However, these classifiers only capture the class-level structures of the base classes.
COMET~\cite{cao2020concept} proposes to generate the embedding features as the combination of the outputs from the concept learners, which highly depend on external knowledge. Instead, our method seeks to discover the shared-latent information without \dy{the }requirements of such knowledge.
\begin{figure*}[!hbtp]
	\centering
	\includegraphics[width=0.9\textwidth]{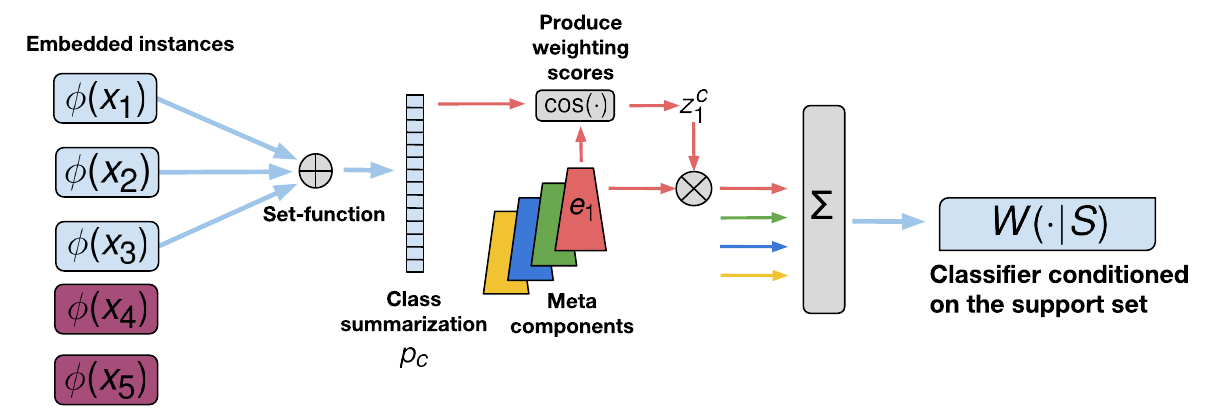}
	\caption{Overview of our Meta Components Learning (MCL) in classifications.  $\phi(x)$ is the instance encoder network. $\{x_1\,x_2,x_3\}$ belongs to the class $c$. We use a set-function to aggregate one class's information. Then, by combining each meta-component with the scores, MCL generates the classifiers for the class $c$. (hint: better view this figure in color mode)} 
	\label{fig:MC}
\end{figure*}

~\\
\noindent\textbf{Optimization-based methods.} Optimization-based appr- oaches~\cite{finn2017model,raghu2019rapid,rusu2018meta,triantafillou2019meta,xu2019metafun} learn a good initialization such that it can be fast adapted to novel tasks based on a few instances. The representative method, MAML~\cite{finn2017model}, learns a network parameter initialization so that the task-specific fine-tuning on the network is more efficient. LEO~\cite{rusu2018meta} learns a parameter generative distribution and adapts the latent code instead of the network parameters in inner loops. MetaFun~\cite{xu2019metafun} learns to summarise a task using a functional representation constructed via iterative updates. 
To enable fast adaptations to any class, the initialization of the classifier in optimization-based methods have to be neutral and cannot preserve any information from prior data. 


~\\

\noindent \textbf{Ensemble learning.} Ensemble learning~\cite{zhou2009ensemble} is similar to our proposed method seemingly but the working principles between these two are different. Ensemble learning~\cite{zhou2009ensemble} is a machine learning paradigm where multiple independent learners are trained to solve the same task. Instead, each meta-component in our proposed method is encouraged to deal with one sub-problem and learn\dy{learns} diverse substructures via episodic training. Additionally, ensemble methods in deep learning normally train a set of neural networks, each as an independent predictive model. But our method shares only one neural network as backbone and the head predictive model is generated by the combined meta-components.

\section{Meta Component Learning}
In this section, we will first introduce the preliminaries about the settings in meta-learning, then introduce the proposed approach: Meta Components Learning (MCL) and Adapted Meta-Component Learning (AMCL).

\subsection{Preliminary: Meta-Learning Framework for Few-shot Learning}

In meta-learning, the model is trained on episodes, which are designed to mimic the few-shot learning tasks. In every episode, a few-shot learning task is constructed as $\{\{(\boldsymbol{x}_i,\boldsymbol{y}_i)\}_{i\in \emph{S}},\{(\boldsymbol{x}_j,\boldsymbol{y}_j)\}_{j\in \emph{Q}}\}$, where $\boldsymbol{x}_i,\boldsymbol{x}_j \in \emph{X}$ are inputs, $\boldsymbol{y}_i,\boldsymbol{y}_j \in \emph{Y}$ supervised information, $\emph{S}$ the support set, and $\emph{Q}$ the query set. The objective is to train a predictive model $f$ conditioned on $\emph{S}$ to achieve minimal expected loss on the query set $\emph{Q}$ as follows
\begin{equation}
\small
    \min_{\phi, \boldsymbol{W}} \; \ell_\emph{Q} (\boldsymbol{y}, f(\phi(\boldsymbol{x}), \boldsymbol{W})),
    \label{meta-train}
\end{equation} 
where $f(\cdot, \boldsymbol{W})$ is the prediction model ($f$ is parameterized by $\boldsymbol{W}$), $\phi$ an encoder network which extracts the feature of the input $\boldsymbol{x}$, and $\ell_\emph{Q}$ the loss function estimated on the query set. Similarly, we denote the loss estimated on support set by $\ell_\emph{S}$.

\subsection{Class Summarization}



A high-quality classifier should be constructed by fully exploring the specific information in that class. Prototypical networks (ProtoNet)~\cite{snell2017prototypical} proposed to consider the mean vector of all embedded support instances in a class, which is called ``prototype'', as the summarization of the instances belonging to the same class.
In practice, we can use any set operator~\cite{ye2020few, zaheer2017deep}, more than the mean operator, to perform class summarization, achieving a permutation invariant embedding of the set of support instances in the class. Set operators include mean pooling, max pooling, min pooling or attention, which are introduced in details in \cite{ye2020few,zaheer2017deep}. With set functions, the class summarization $\boldsymbol{p}_c$ is obtained as
\begin{equation}
\small
    \boldsymbol{p}_c = \text{SetFunction}(\{\phi(\boldsymbol{x}_i)\}_{i\in \emph{S}_c}),
    \label{Pooling}
\end{equation}
where $\emph{S}_c$ denotes the support set of class $c$.
By choosing an appropriate set function concerning the learning task, the class information can be well-extracted. Then by exploring the correlation between the class information and the meta-components, a high-quality classifier is constructed.

\subsection{Meta Components}



We propose a novel design---the meta-component, to construct a robust classifier for an unseen task. In the task of visual classifications, we could depict any category image as a linear combination of visual attributes~\cite{changpinyo2016synthesized}. For example, an image of cat could be represented as: $0.3\cdot$long beards + $0.2\cdot$short fur$ - 0.2\cdot$long legs. Likewise, an unseen class's prediction model could also be replaced by a weighted linear combination of visual attributes' predictors. 
Inspired by this example, we propose to construct a classifier by a weighted combination of meta-components, denoted by $\boldsymbol{E}=[\boldsymbol{e}_1, \boldsymbol{e}_2, \ldots, \boldsymbol{e}_N]^T$, where $N$ is the number of meta-components. Each vector $\boldsymbol{e}_n$ is expected to learn to classify a subclass-level structure. The subclass-level structure can be an attribute or a concept in a class. However, we do not assume any prior knowledge about the supervision information of the attributes or concepts. Instead, we learn the meta-component to capture them along with meta-episodes. 


What remains is to determine the combination scores to combine the meta-components for a classifier. In the proposed approach, we define the combination scores as the similarity between the class summarization $\boldsymbol{p}_c$ and the meta-components. To be specific, the combination scores between the $c$-th class and the $n$-th meta-component, denoted by $z_n^c$, is computed by the cosine similarity as
\begin{equation}
\small
    z_n^c = \text{cos}(\boldsymbol{p}_c,\boldsymbol{e}_n)=\frac{\boldsymbol{p}_c}{\|\boldsymbol{p}_c\|}\cdot\frac{\boldsymbol{e}_n}{\|\boldsymbol{e}_n\|}.
    \label{cos sim}
\end{equation}

Then the classifier of the $c$-th class is parameterized as
\begin{equation}
\small
    \boldsymbol{w}_c = \sum_{n=1}^N z_n^c \boldsymbol{e}_n.
    \label{eqn:construct classifier}
\end{equation}

The parameter $\boldsymbol{W}$ in the predictive model $f$, is obtained as $\boldsymbol{W} = [\boldsymbol{w}_1, \boldsymbol{w}_2, ..., \boldsymbol{w}_{N_c}]$, where $N_c$ is the number of classes sampled in one episode. In classification tasks, the predictive model $f$ is set as a fully-connected layer (a linear classification model) but without bias. As $\boldsymbol{W}$ is a function of $\boldsymbol{E}$, the gradient of $\boldsymbol{W}$ can be back-propagated to $\boldsymbol{E}$ and $\boldsymbol{E}$ can be optimized by minimizing the loss function in Eqn.~\eqref{meta-train}.

\kai{The proposed method, named Meta Component Learning (MCL) is summarized in Algorithm 1, which is available in Appendix.}

\subsection{Score Adaptations}

In the proposed approach, the \dy{combination }scores are initialized as the similarity between the class summarization and the meta-components. As the classifier is combined with the weights being combination scores, its performance depends on the scores. 
To further improve the performance of the classifiers in a novel task, we propose to adapt the combination scores on the support set \emph{S} by performing a few steps of gradient descent as follows, 
\begin{eqnarray}
\small
    & \boldsymbol{\zeta}^{(i+1)} = \boldsymbol{\zeta}^{(i)} - \alpha \cdot \nabla_{\boldsymbol{\zeta}^{(i)}}\ell_\emph{S}(\boldsymbol{y}, f(\phi(\boldsymbol{x}), \boldsymbol{E}^\top \boldsymbol{\zeta}^{(i)})), \nonumber
    \\
    & \qquad \qquad\qquad\qquad\text{for } i=1,\ldots,M
     \label{eqn:adapt}
\end{eqnarray}
\dy{where $\boldsymbol{\zeta}$ is the score matrix whose $(n,c)$ entry is the combination score $z_n^c$.} Here $M$ is the number of adaptation steps. Note that the encoder $\phi$ and the meta-component $\boldsymbol{E}$ are fixed during this adaptation, and are updated by minimizing the loss on the query set. Unlike optimization-based meta-learning, we do not aim to learn the initialization of $\boldsymbol{\zeta}$. Instead, we initialize $\boldsymbol{\zeta}$ by Eqn.~\eqref{cos sim}. This reduces the complexity of the optimization problem. The proposed approach is named Adapted Meta-Component Learning (AMCL), and summarized in Algorithm 2 in Appendix.

\subsection{Disentangling Regularizer}


To discover diverse substructures, we apply a disentangling regularizer on the meta-components. If there is no constraint on the meta-components, they may be entangled with each other. If so, the meta-components may only focus on the dominant substructures, and thus, fail to capture the infrequent ones. As an infrequent component in the seen tasks may become an important component in a novel task, to improve the generalization and avoid overfitting to the seen tasks, it is important to encourage the learning to capture the infrequent components by disentangling the meta-components. Inspired by~\cite{brock2018large, xie2018orthogonality}, we apply an orthogonality-promoting regularizer on the meta-components as
\begin{equation}
\small
    R(\boldsymbol{E})=\|(\boldsymbol{E}^\top \boldsymbol{E})^ {\circ (\boldsymbol{1}-\boldsymbol{I})} \|^2 = \sum_{ij} {(\boldsymbol{E}^\top \boldsymbol{E})}_{ij}^{{(\boldsymbol{1}-\boldsymbol{I})}_{ij}},
    \label{ortho_reg_2}
\end{equation}
where $\boldsymbol{A}^{\circ \boldsymbol{B}}$ is the element-wise power, returning a matrix raising each element of $\boldsymbol{A}$ to the corresponding powers in $\boldsymbol{B}$, $\boldsymbol{1}$ is an all-ones matrix, and $\boldsymbol{I}$ is an identity matrix.  Our applied regularizer takes similar effects with the one in BigGAN~\cite{brock2018large}, as they both achieve minimal value when $\boldsymbol{E}^\top \boldsymbol{E} = \boldsymbol{I}$. Empirically, the regularizer proposed in Eqn.~\eqref{ortho_reg_2} performs better on the tasks of this study than the regularizer in BigGAN. 

With the orthogonality-promoting regularizer, the objecive function becomes
\begin{equation}
\small
    \min_{\phi, \boldsymbol{E}} \;  \ell_\emph{Q}(\boldsymbol{y}, f(\phi(\boldsymbol{x}), \boldsymbol{E}^\top \boldsymbol{\zeta})) + \beta  R(\boldsymbol{E}),
    \label{obj}
\end{equation}
where $\beta$ is a regularizer parameter controlling the effect of the orthogonality-promoting regularizer in the loss function.

\section{MCL and AMCL for Regression and Reinforcement Learning}

In the previous section, we present MCL and AMCL for classification problems. In this section, we will show how to extend the proposed method for the regression tasks and reinforcement learning tasks. 


\subsection{Regression}
In few-shot regression tasks, a model is used to fit a sinusoidal curve with limited labeled instances. Having x-axis coordinates $x_i$ alone can not represent the full characteristics of a task. A context encoder $\phi'$ is used to encode $\{[x_i,y_i]\}_{i\in \emph{S}}$ for constructing a proper task summarization, where $[\cdot]$ is concatenation. The task summarization $\boldsymbol{p}$ is obtained as
\begin{equation}
\small
    \boldsymbol{p} = \text{SetFunction}(\{\phi'([x_i,y_i])\}_{i\in \emph{S}}).
    \label{Pooling-Regression}
\end{equation}

\kai{
In regression tasks, there is no class summarization. Instead, we use the task summarization for combination codes prediction. However, the dimension of task summarization and meta components are different, making it impossible to directly compute the cosine similarity between them. To address this issue, we assign a score prediction vector, denoted by $\boldsymbol{\theta}_n$, for each component. These vectors are with the same dimension as the task summarization. Thus, the combination scores for each component can be computed by
\begin{equation}
\small
    z_n = \frac{\boldsymbol{p}}{\|\boldsymbol{p}\|}\cdot\frac{\boldsymbol{\theta}_n}{\|\boldsymbol{\theta}_n\|}.
    \label{Regression Scores}
\end{equation}

The score prediction vectors are meta-trained together with the meta-components $\boldsymbol{E}$ by minimizing the few-shot tasks' loss on query set. Then the \dy{predictive} model is computed via Eqn~\eqref{eqn:construct classifier}. }







\subsection{Reinforcement Learning}
In reinforcement learning (RL), we consider a Markov decision process (MDP) as a task. MCL or AMCL receives a sequence of observation-action-reward tuples $(\boldsymbol{o}_t,\boldsymbol{a}_t,r_t)$, where $\boldsymbol{o}_t$ is the observation at step $t$, $\boldsymbol{a}_t$ the action, and $r_t$ the reward received at step $t$. 


The action is sampled $\boldsymbol{a}_t \sim f_{\pi}(\boldsymbol{a}_t|\boldsymbol{o}_t,\boldsymbol{s}_t)=f_{\pi}(\phi(\boldsymbol{o}_t), \boldsymbol{W})$, where the policy network $f_{\pi}$ is parameterized by $\boldsymbol{W}$ and a feature extractor $\phi$ (an encoder network). The task summarization is built with the below equation:
\begin{equation}
\small
    \boldsymbol{p} = \text{SetFunction}(\{\phi'([\boldsymbol{o}_i, \boldsymbol{s}_i, r_i])\}_{i\in \emph{S}}),
    \label{Pooling-rl}
\end{equation}
where the context encoder $\phi'$ encode the information for each $\{[\boldsymbol{o}_i, \boldsymbol{s}_i, r_i]\}$ in the support $\emph{S}$. 
\kai{Similar to the case of regression tasks, we set an additional set of vectors $\{\boldsymbol{\theta}_n\}_{n=1}^{N}$} for combination score prediction as in Eqn \eqref{Regression Scores}. 


In each task, we first sample $K$ support rollouts $\emph{S}$ for constructing the predictive model $f_{\pi}$ and the task-specific policy network minimizes the loss on another $K$ query rollouts $\emph{Q}$ as
\begin{multline}
\small
        \min_{\phi,\phi',\boldsymbol{W}} \ell_Q(f_{\pi}(\phi'(\boldsymbol{o}_j), \boldsymbol{W}))\\ 
    =\min_{\phi,\phi',\boldsymbol{W}} -\mathbb{E}_{q,q_0,f_{\pi}}[\sum_{j} \gamma^j r(\boldsymbol{s}_j,\boldsymbol{a}_j)],
    \label{rl-loss}
\end{multline}
where $\gamma$ is a discount factor, $q$ is the transition probability distribution, and $q_0$ is the initial state distribution. 

\section{Experiments}

In this section, we empirically evaluate the proposed approaches MCL and AMCL on the few-shot classification problem\dy{problems} on \emph{mini}ImageNet, \emph{tiered}ImageNet, FS-CIFAR100 and CUB datasets. Besides, we also verify the performance on regression and reinforcement learning problems, showing the practicability of the proposed method on problems other than classification. 

\subsection{Few-shot Classification}
\begin{table*}[ht]
	\centering
	\resizebox{1\columnwidth}{!}{
		\begin{tabular}{c c c c c}
		    \toprule[1pt]
			 \multirow{2}{*}{\textbf{model}}& \multicolumn{2}{c}{\textbf{1-Shot}} & \multicolumn{2}{c}{\textbf{5-Shot}}  \\ \cline{2-3} \cline{4-5}
			 & \textbf{ConvNet} & \textbf{ResNet} & \textbf{ConvNet} & \textbf{ResNet} \\
			\hline
			Matching Networks~\cite{vinyals2016matching} & 43.56 $\pm$ 0.84 & - & 55.31 $\pm$ 0.73 & - \\
			MAML~\cite{finn2017model} & 48.70 $\pm$ 1.84 & - & 63.15 $\pm$ 0.91  & - \\
			Prototypical Networks~\cite{snell2017prototypical} & 49.42 $\pm$ 0.78 & - &  68.20 $\pm$ 0.66  & -\\
			Relation Networks~\cite{sung2018learning} & 50.44 $\pm$ 0.82& - & 65.32 $\pm$ 0.70 & - \\
			SNAIL~\cite{mishra2017simple} & - & 55.71 $\pm$ 0.99 & - & 68.88  $\pm$ 0.92  \\
			TADAM~\cite{oreshkin2018tadam} & - & 58.50 $\pm$ 0.30& -& 76.70 $\pm$ 0.30  \\
			LEO~\cite{rusu2018meta} & - & 61.76 $\pm$ 0.08 & - & 77.59 $\pm$ 0.12 \\
			MetaOptNet~\cite{lee2019meta} & - & 62.64 $\pm$ 0.61 & - & 78.63 $\pm$ 0.46 \\	
			TapNet~\cite{yoon2019tapnet} & 50.68 $\pm$ 0.11 & 61.65 $\pm$ 0.15 & 69.00 $\pm$ 0.09 & 76.36 $\pm$ 0.10\\
			DSN~\cite{simon2020adaptive} & - & 62.64 $\pm$ 0.66 & - & 78.83 $\pm$ 0.45\\
			CAN~\cite{hou2019cross} & - & 63.85 $\pm$ 0.48 & - & 79.44 $\pm$ 0.34\\
			FEAT~\cite{ye2020few} & 55.15 $\pm$ n/a& 66.78 $\pm$ n/a & 71.61 $\pm$ n/a & 82.05 $\pm$ n/a \\
			\midrule[1pt]
            MCL & 55.31 $\pm$ 0.20 & 65.86 $\pm$ 0.20 & 71.93 $\pm$ 0.16 & 82.05$\pm$ 0.13 \\
            AMCL & \textbf{55.35 $\pm$ 0.20} & \textbf{66.92 $\pm$ 0.20} & \textbf{72.37 $\pm$ 0.16}  & \textbf{82.47$\pm$ 0.13} \\
			\bottomrule[1pt]
	\end{tabular}}
	\caption{Experimental results with a $95\%$ confidence interval on \emph{mini}ImageNet. (n/a means ``not available'')}
	\label{table:ImageNet}
\end{table*}
\subsubsection{Implementation Details}

Following \cite{chen2019closer,chen2020new,gidaris2018dynamic}, we adopt two training stages: pre-training and meta-training stage. In the pre-training stage, we train the feature extractor $\phi$, which is parameterized by a deep CNN backbone. The training loss is a classification loss, defined by the cross-entropy loss , on all seen classes with a linear classifier $f(\cdot|\boldsymbol{W}_b)$, where $\boldsymbol{W}_b \in \mathbb{R}^{d\times N_c}$, $d$ is the dimension of embedded instances and $N_c$ is the number of seen classes.


The ConvNet-64 and ResNet-12 are selected as backbones as they are widely applied in other meta-learning approaches. In practice, we pre-train the feature extractor on meta-training set for $500$ epochs with a batch size being $16$ and an SGD optimizer with an initial learning rate being $10^{-3}$. Besides, data augmentation, including random crop, left-right flip, and color jitters, is applied. Note that the data augmentation will not be applied in the meta-training stage.

In the second stage, the meta-training stage, we train the meta-learning models, including the feature extractor and the meta-components, with Eqn.~\eqref{meta-train}. The model is trained on $20000$ episodes and the number of query instances in each class is set to $15$ for both training and testing. We adopt \dy{an} SGD optimizer with an initial learning rate being $0.002$. As the backbone is initialized with the pre-trained weights, its learning rate is scaled by $0.1$.

We follow ProtoNet~\cite{snell2017prototypical} to use a mean-pooling as the set-function in Eqn.~\eqref{Pooling} in classification tasks. As for the final results, they are reported as mean accuracy with a $95\%$ confidence interval over 10,000 episodes, following FEAT~\cite{ye2020few}. 

\subsubsection{Datasets}
\label{datasets}
We conducted experiments on four benchmark datasets: \emph{mini}ImageNet, \emph{tiered}ImagetNet, Fewshot-CIFAR100 (FS-FC100) and Caltech-UCSD Birds-200-2011 (CUB). 

\textbf{\emph{mini}ImageNet}~\cite{vinyals2016matching} includes $100$ randomly chosen classes from ILSVRC-2012~\cite{russakovsky2015imagenet}. These classes are split into $64$, $16$ and $20$ classes for meta-training, meta-validation and meta-testing. Each class contains $600$ size $84 \times 84$ images and the total images are $60,000$ images. 

\textbf{\emph{tiered}ImageNet}~\cite{ren2018meta} is a larger subset of ILSVRC-2012, consisted of $608$ classes grouped into $34$ high-level categories. They are divided into $20$ categories for meta-training, $6$ categories for meta-validation, and $8$ categories for meta-testing. This corresponds to $351$, $97$, and $160$ classes for meta-training, meta-validation, and meta-testing, respectively. \kai{Note that neither high-level categories nor low-level categories, i.e. classes, are shared in meta-training and meta-testing set.} The size of images is $84 \times 84$.

\textbf{Fewshot-CIFAR100}~\cite{oreshkin2018tadam} is a few shot image-based dataset based on CIFAR100~\cite{krizhevsky2009learning}. It is also referred to as FC100. FC100 consists of $32\times 32$ color images belonging to $100$ different classes, $600$ images per class. The $100$ classes are further grouped into $20$ superclasses: $60$ classes from $12$ superclasses for meta-training, $20$ classes from $5$ superclasses for meta-validation, and meta-testing separately.
\begin{table}[h]
	\centering
	\resizebox{0.9\columnwidth}{!}{
		\begin{tabular}{c c c}
		    \toprule[1pt]
			\textbf{model}   & \textbf{1-Shot}& \textbf{5-Shot}  \\
			\hline
			cosine classifier~\cite{chen2019closer}& 61.49 $\pm$ 0.91  & 82.37 $\pm$ 0.67\\
			LEO~\cite{rusu2018meta}	&66.33 $\pm$ 0.05	&81.44 $\pm$ 0.09\\
			Prototypical Networks~\cite{snell2017prototypical} &	65.65 $\pm$ 0.92&	83.40 $\pm$ 0.65\\
			MetaOptNet~\cite{lee2019meta}  & 65.99 $\pm$ 0.72 &	81.56 $\pm$ 0.63\\
			DSN~\cite{simon2020adaptive} & 66.22 $\pm$ 0.75 & 82.79 $\pm$ 0.48 \\
			CTM~\cite{li2019finding} 	& 68.41 $\pm$ 0.39 & 84.28 $\pm$ 1.73\\
			SimpleShot~\cite{wang2019simpleshot}  & 69.09 $\pm$ 0.22 & 84.58 $\pm$ 0.16 \\
			CAN~\cite{hou2019cross} & 69.89 $\pm$ 0.51 & 84.23 $\pm$ 0.37\\
			FEAT~\cite{ye2020few}  &	70.80 $\pm$ 0.23 &	84.79 $\pm$ 0.16 \\	
			\midrule[1pt]
            MCL &   70.84 $\pm$ 0.23  &	85.65 $\pm$ 0.15 \\
            AMCL  & \textbf{71.00 $\pm$ 0.22}  &	\textbf{85.67 $\pm$ 0.15}  \\

			\bottomrule[1pt]
	\end{tabular}}
	\caption{Experimental Results\dy{results} with a $95\%$ confidence interval on \emph{tiered}ImageNet. LEO's backbone is $\emph{WRN-28-10}$. CTM and SimpleShot use the $\emph{ResNet18}$ backbone. The others use $\emph{ResNet12}$.}
	\label{table:TieredImageNet}
\end{table}

\textbf{Caltech-UCSD Birds (CUB) 200-2011}~\cite{wah2011caltech} The CUB dataset contains in total $11,788$ images of birds from $200$ species. We follow the settings in FEAT~\cite{ye2020few} to randomly sample $100$ species for meta-training, $50$ species for meta-validation, and use the remaining $50$ species for meta-testing. 

\subsubsection{Main Results}
We compare our proposed MCL and AMCL with various state-of-the-art algorithms on 1-shot 5-way and 5-shot 5-way  classification tasks.

The results on \emph{mini}ImageNet, \emph{tiered}ImageNet, CUB and Fewshot-CIFAR100 are shown in Tables \ref{table:ImageNet}, \ref{table:TieredImageNet}, \ref{table:CUB}, and \ref{table:FS-FC100}. Note that the results of baseline algorithms are the reported value in the original papers. From Table \ref{table:ImageNet}, we first observe AMCL achieves the highest accuracies than the others for both 1-shot and 5-shot settings using different backbones on \emph{mini}ImageNet, indicating the effectiveness of the proposed method. MCL also performs well but is slightly worse than AMCL. This verifies the validity of score adaptation. The same observation is noticed on \emph{tiered}ImageNet in Table \ref{table:TieredImageNet}, where both MCL and AMCL outperform the baseline methods. And the best results are achieved by AMCL. Although on CUB dataset (Table \ref{table:CUB}), AMCL performs worse than MCL and Matching Networks in 1-shot experiments, it still gets significant improvement over Matching Networks in 5-shot experiments, achieving the 87.18\%. In Fewshot-CIFAR100 dataset, as observed in Table \ref{table:FS-FC100}, the accuraci of both MCL and AMCL are significantly higher than other baselines in both 1-shot and 5-shot experiments. Note that in \cite{sun2019meta} MTL achieves higher accuracies for Fewshot-CIFAR100 with a special meta-training paradigm---hard task meta-batch. Since our method still follows a normal meta-training manner, we only report their results under this same setting. 
As the proposed approach achieves the highest accuracy in all the datasets, its effectiveness is verified.
\begin{table}[h]
	\centering
	\resizebox{0.9\columnwidth}{!}{
		\begin{tabular}{c  c c}
		    \toprule[1pt]
			\textbf{model}   & \textbf{1-Shot}& \textbf{5-Shot}  \\ 
			\hline
			Prototypical Networks~\cite{snell2017prototypical}  &	66.09 $\pm$ 0.92 &	82.50 $\pm$ 0.58\\
			RelationNet~\cite{chen2019closer,sung2018learning} & 66.20 $\pm$ 0.99 & 82.30 $\pm$ 0.58\\
			MAML~\cite{chen2019closer,finn2017model}  &	67.28 $\pm$ 1.08 &	83.47 $\pm$ 0.59\\
			cosine classifier~\cite{chen2019closer}  &	67.30 $\pm$ 0.86 & 84.75 $\pm$ 0.60\\
			Matching Networks~\cite{vinyals2016matching}  &	71.87 $\pm$ 0.85 & 85.08 $\pm$ 0.57\\
			\midrule[1pt]
            MCL  & \textbf{74.94  $\pm$ 0.22} &	86.98 $\pm$ 0.13 \\
            AMCL  & 74.81 $\pm$ 0.21 & \textbf{87.18 $\pm$ 0.12} \\
			\bottomrule[1pt]
	\end{tabular}}
	\caption{Experimental results with a $95\%$ confidence interval on CUB. RelationNet and MAML use the \emph{ResNet34} backbone. The others use the \emph{ResNet12} backbone.}
	\label{table:CUB}
\end{table}

\begin{table}[h]
	\centering
	\resizebox{0.9\columnwidth}{!}{
		\begin{tabular}{c c c c }
		    \toprule[1pt]
			\textbf{model}   & \textbf{1-Shot}& \textbf{5-Shot}  \\ 
			\hline
			cosine classifier~\cite{chen2019closer}   &	38.47 $\pm$ 0.70&	57.67 $\pm$ 0.77\\
			TADAM~\cite{oreshkin2018tadam}  &	40.10 $\pm$ 0.40 &	56.10 $\pm$ 0.40\\
			MetaOptNet~\cite{lee2019meta}  & 41.10 $\pm$ 0.40 &	55.50 $\pm$ 0.60\\
			Prototypical Networks~\cite{snell2017prototypical}  & 41.54 $\pm$ 0.76 & 57.08 $\pm$ 0.76\\
			MTL~\cite{sun2019meta}   &	43.60 $\pm$ 1.80   &	55.40 $\pm$ 0.90\\
		    Matching Networks~\cite{vinyals2016matching}  &	43.88 $\pm$ 0.75 &	57.05 $\pm$ 0.71\\
			\midrule[1pt]
            MCL &  44.74 $\pm$ 0.18 &	\textbf{61.18 $\pm$ 0.18} \\
            AMCL  & \textbf{45.05 $\pm$ 0.19} & 60.89 $\pm$ 0.18 \\
			\bottomrule[1pt]
	\end{tabular}}
	\caption{Experimental results with a $95\%$ confidence interval on FS-FC100. All methods apply the \emph{ResNet12} backbone.}
	\label{table:FS-FC100}
\end{table}

\subsection{Ablation Study}
\subsubsection{Effects of Orthogonality-Promoting Regularizer}
To learn diverse meta-components, an orthogonality-promoting regularizer is applied on the meta-components. By changing the regularizer parameter $\beta$ in Eqn.~\eqref{ortho_reg_2}, we explore the effect of the regularizer on the performance of the proposed model. The results in Table \ref{table:Adapt beta} show that with appropriate $\beta$, AMCL gains 0.88\% higher accuracy in the 5-way-1shot task and 0.66\% higher accuracy in 5-way-5-shot tasks over the model learned without the regularizer. This verifies the benefit of the orthogonality-promoting regularizer.

\begin{table}
	\centering
	\resizebox{0.7\columnwidth}{!}{
		\begin{tabular}{c c c }
		    \toprule[1pt]
			\textbf{Setups}  & \textbf{1-Shot}& \textbf{5-Shot}  \\ 
			\hline
			$\beta = 0$ & 70.12 $\pm$ 0.23 & 85.01 $\pm$ 0.16 \\
			$\beta =0.0025$ &	70.43 $\pm$ 0.23 &85.43 $\pm$ 0.15\\
			$\beta =0.25$ &	70.94 $\pm$ 0.23 &	85.52 $\pm$ 0.15\\
			$\beta =0.5$ & \textbf{71.00 $\pm$ 0.22} & \textbf{85.67 $\pm$ 0.15}\\
			$\beta =2.5$ & 70.97 $\pm$ 0.23  & 85.58$\pm$ 0.15\\
			\bottomrule[1pt]
	\end{tabular}}
	\caption{The effect of orthogonality promoting regularizer on \emph{tiered}ImageNet's performance with the ResNet-12 backbone}
	\label{table:Adapt beta}
\end{table}
\subsubsection{Influence of Number of Meta Components}
To explore the relationship between the performance and the number of meta-components, we conduct experiments on \emph{tiered}ImageNet by varying number of meta-components from $160$ to $1,280$. To avoid the effect of adaptation, we conduct the experiment on MCL. Table \ref{table:Nun MetaFeat} shows that when the number of meta-components equals the hidden dimension of the embedded instances $\phi(x)$, which equals to $640$, MCL achieves the best performance. \kai{When $N \le 640$, the performance increases for a larger $N$ as more information can be captured with more components. However, when the number of components exceeds the dimension of the embedded features, no extra information will be captured with more components, and the increases of number of combination scores makes it difficult to predict the optimal scores for each component. Thus, the performance deteriorates with the increase of $N$. }

\begin{table}[h]
	\centering
	\resizebox{0.7\columnwidth}{!}{
		\begin{tabular}{c c c }
		    \toprule[1pt]
			\textbf{Setups}  & \textbf{1-Shot}& \textbf{5-Shot}  \\ 
			\hline
			$N=160$ & 69.17 $\pm$ 0.23 & 83.77 $\pm$ 0.16  \\
			$N=320$ &	69.64 $\pm$ 0.23 &	84.38 + 0.16\\
			$N=640$ &	\textbf{70.79 $\pm$ 0.23}  &	\textbf{85.65 $\pm$ 0.15}\\
			$N=960$ & 70.12 $\pm$ 0.16 & 84.71 $\pm$ 0.16 \\
			$N=1280$ & 70.27 $\pm$ 0.23  &  84.77 $\pm$ 0.16 \\
			\bottomrule[1pt]
	\end{tabular}}
	\caption{The effect of number of meta-components on \emph{tiered}ImageNet's performance with ResNet-12 backbone.}
	\label{table:Nun MetaFeat}
\end{table}

\subsection{What do Meta-components Learn?}
\label{What mc learn}
\begin{figure}[h]
	\centering
	\resizebox{0.7\columnwidth}{!}{
		\includegraphics[width=0.4\columnwidth]{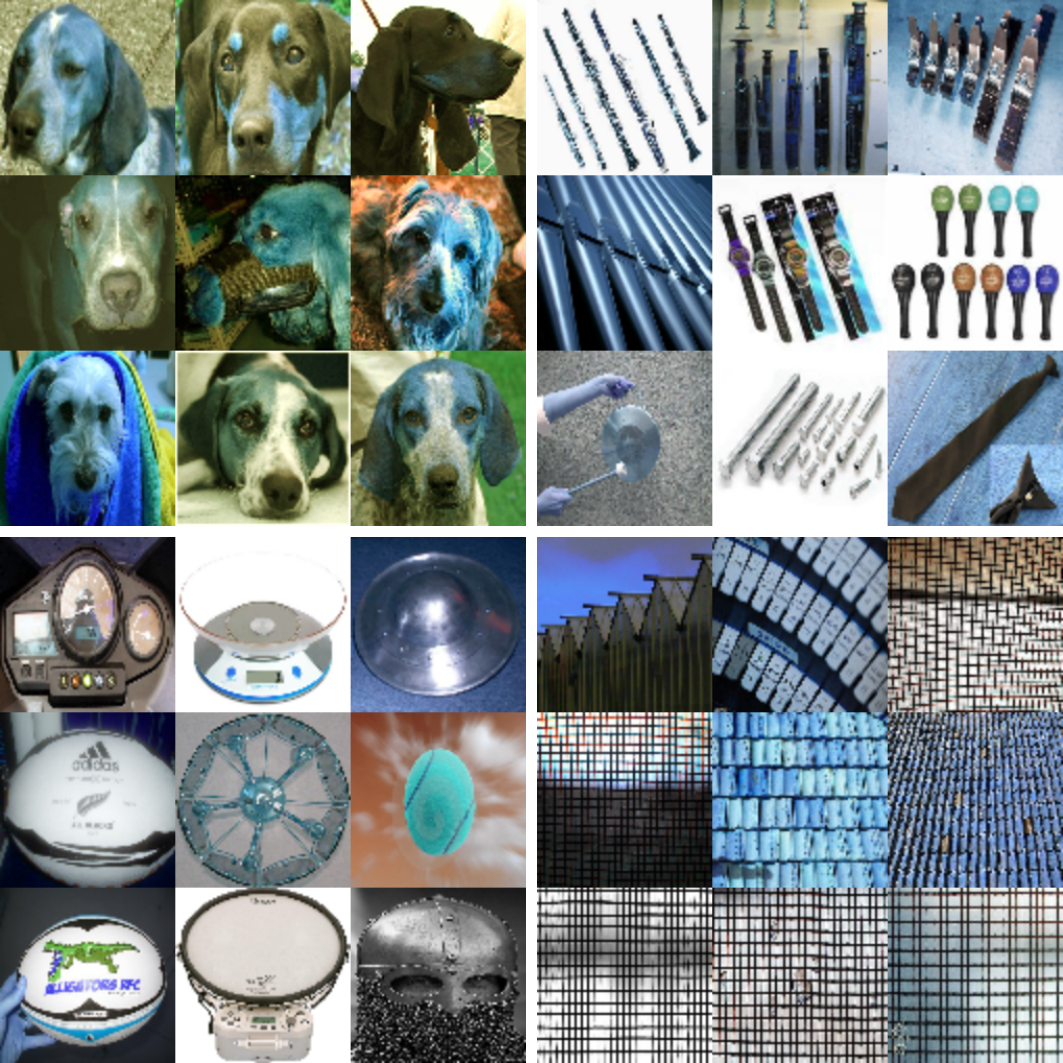}
	}
    \caption{Four groups of images from \emph{Tiered}ImageNet with the highest combination scores for a particular meta-component}
    \label{fig:mc_learned}
\end{figure}

In this section, we test whether the meta-components can discover subclass-level structures as proposed. To visualize the subclass-level structure(s) a meta-component captures, we display the images that have the largest similarities with this component as in Fig.~\ref{fig:mc_learned}.



The meta-components have discovered diverse latent substructures. As shown in Fig.~\ref{fig:mc_learned}, the meta-component for the upper-left corner corresponds to the dog head, the one for the upper-right corner the stripe-shaped structure, the one for the bottom-left corner an oval structure, and the one for the bottom-right corner crowded-straight lines.


To qualitatively analyze the meta-components, we build a small toy-dataset. The dataset consists of colored geometric patterns. The geometric shapes include circle, triangle, quadrilateral, hexagon and pentagon. The colors include black, blue, green, orange, red and yellow. 

We collect the meta-trained meta-components and the batch-trained attribute classifiers for five shapes and six colors. The number of meta-components is set to $11$. We calculate the Pearson Correlation Coefficients between meta-components and the attribute classifiers.

\begin{figure}[h]
	\centering
	\resizebox{1\columnwidth}{!}{
		\includegraphics[width=0.4\columnwidth]{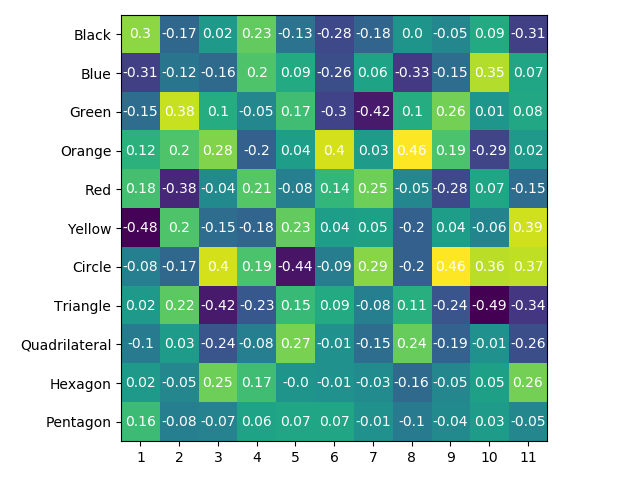}
	}
\caption{Pearson Correlation Coefficients between meta-components and attribute learners (classifiers).}
\label{fig:mc_heat}
\end{figure}

As shown in Fig.~\ref{fig:mc_heat}, the x-axis corresponds to all $11$ meta-components. Although we train the attribute learners and meta-components separately, they still have a strong correlation. The meta-components can be regarded as linear combinations of the attribute learners. Since these attributes are perceptible substructures, we could conclude that our meta-components could learn the substructures as we claimed in the section~\ref{introduction}.

\subsection{Regressions}

We follow MAML\cite{finn2017model} to carry out experiments on sinusoidal regression tasks. Each task contains the input and output of a sine wave, which is specified by the amplitude and phase of the sinusoid. The amplitude varies between $[0.1, 5.0]$ and the phase varies between $[0, \pi]$. During training, $5$ labeled datapoints (randomly sampled from $[-5, 5]$) for $5$-shot and $10$ labeled datapoints for 10-shot are given in each task as support set, and $15$ data points with only inputs are given as query set. During the evaluation, we generate $1000$ new tasks, each task having $1000$ points (evenly spaced samples over the interval $[-5, 5]$). We randomly choose 5 points or 10 points as a support set and set the rest as a query set. The loss function is a mean-squared error (MSE). 
\begin{figure}[h]
	\centering
	\resizebox{1.05\columnwidth}{!}{
			\centering
			\includegraphics[width=0.4\columnwidth]{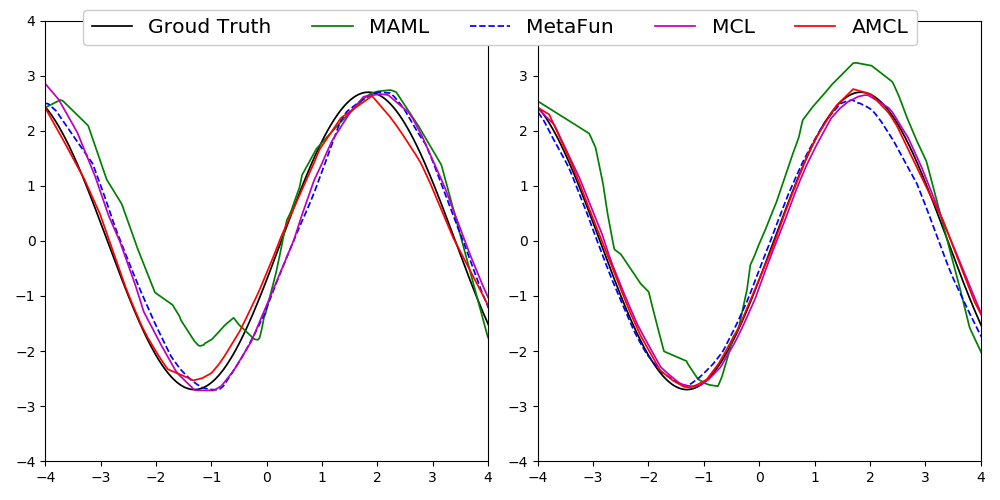}
			\label{fig:regression}}
\caption{Left: 5-shot regression. Right: 10-shot regression. AMCL fits  the original sine curve better than other methods.}
\label{fig:prototype-test}
\end{figure}
For a fair comparison, the regressor is set to be a small neural network with an input layer of size $1$, followed by $2$ hidden layers of size $40$ with ReLU nonlinear activations, and an output layer of size $1$. The results averaged over the sampled 1000 sine curves with 95\% confidence intervals are presented in table~\ref{table:regression}.
\begin{table}[h]
	\centering
	\resizebox{0.9\columnwidth}{!}{
		\begin{tabular}{c c c}
		    \toprule[1pt]
			\textbf{model}   & \textbf{5-Shot MSE}& \textbf{10-Shot MSE}  \\ 
			\hline
			MAML~\cite{finn2017model}  &	1.100 $\pm$ 0.077 &	0.648 $\pm$ 0.046\\
			Meta-SGD~\cite{li2017meta} &  0.900 $\pm$ 0.160 & 0.530 $\pm$ 0.090 \\
			CAVIA~\cite{zintgraf2019fast} & - & 0.190 $\pm$ 0.020\\
			MetaFun~\cite{xu2019metafun}	 &  0.076 $\pm$ 0.004 & 0.054 $\pm$ 0.004\\
			\midrule[1pt]
            MCL  & 0.117 $\pm$ 0.014 &	0.051 $\pm$  0.009 \\
            AMCL  & \textbf{0.058 $\pm$ 0.010} & \textbf{0.017 $\pm$ 0.003} \\
			\bottomrule[1pt]
	\end{tabular}}
	\caption{Experimental Results on Sinusoidal Regression. We report MSE results as the mean and $95\%$ confidence interval. }
	\label{table:regression}
\end{table}

From the table~\ref{table:regression}, MCL beats all the methods except for MetaFun on the 5-shot task and AMCL performs better than the other methods with an obvious margin. Notice that the mean results of MetaFun~\cite{xu2019metafun} reported in its paper is $0.040 \pm 0.008$ for 5-shot MSE and $0.017\pm 0.005$ for 10-shot MSE. However, such results are achieved using an MLP with 128 nodes per hidden layer, which is much wider than the backbone network in our experiments. Fig.~\ref{fig:prototype-test} also shows that the curves of our method fits ground truth better, indicting its superiority in predicting the regression values. 
\subsection{Reinforcement Learning}
In this experiment, we evaluate our method on 2D Navigation tasks, following the setting in MAML\cite{finn2017model}. The agent moves in a 2D world using continuous actions, and at each time-step, a negative reward proportional to its distance from a pre-defined goal position is given. A goal position is randomly chosen for each task.
We perform the experiments with the same procedure as~\cite{finn2017model}. Goals are sampled from an interval of $(x, y) = [-0.5, 0.5]$. The observation is the current 2D position. Actions correspond to velocity commands clipped to be in the range $[-0.1, 0.1]$. In meta-training, we sample 20 rollouts as support set and the total reward over 20 rollouts is measured. In evaluations, we sample 20 new unseen tasks with 20 rollouts per task as the support set and 40 rollouts per task to verify the performance. We use a linear value function and a policy network with two hidden layers of size 100 with ReLU nonlinear activation. The policy network is optimized to maximize the total rewards. 

\begin{figure}[!h]
	\centering
	\resizebox{0.9\columnwidth}{!}{
		\includegraphics[width=0.4\columnwidth]{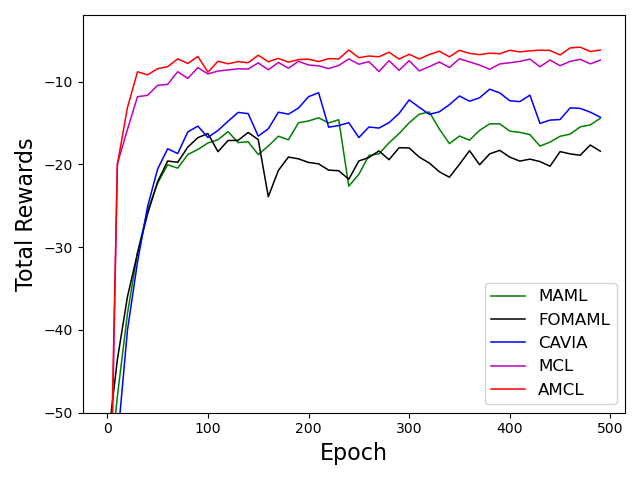}
		\label{fig:rl}
	}
\caption{Total rewards of different methods the in 2D Navigation task.}
\label{fig:rl}
\end{figure}

Figure~\ref{fig:rl} shows the evaluation results. We can observe that both MCL and AMCL outperform the other methods. MCL and AMCL comes to the optimal point very quickly with fewer oscillations. AMCL keeps better than the other methods along the training process and also has a much smoother curve.

\section{Conclusion}

In this paper, we propose a new meta-learning algorithm, called Meta-component Learning (MCL), for few-shot learning. MCL could learn to construct the predictive model as weighted combinations of meta-learned components. The components explore the subclass-level structures, and allow the predictive model in classifications to better explore the class-specific information.To reduce the information redundancy, we apply an orthogonality regularizer and let meta-components learn diverse substructure representations. Through updating the matching score, i.e. the combination scores, AMCL is proposed to adapt the classifiers to a task-specific space and attain high generalization capabilities. We evaluated our algorithm and achieve SOTA performance on four few-shot classification benchmarks, sinusoidal regressions and reinforcement learnings.

\qh{put into appendix?}
Our algorithm consumes very small time complexity. The time complexity of our approach in classifications is $\mathcal{O}(Nd(N_c + M))$, where $N$ is the number of the meta-components, $d$ the dimension of the embeded instances, $N_c$ the number of ways in a few-shot task, and $M$ the number of the adaptation steps ($M=0$ in MCL). Assuming in regressions and reinforcement learning, the width and depth of our context encoder network are bounded by $H$ and $D$, respectively, the time complexity of our approach is $\mathcal{O}(Nd(N_c + M)+HD)$.

{\small
\bibliographystyle{ieee_fullname}
\bibliography{egbib}
}

\newpage
\newpage
\begin{appendices}

\begin{algorithm}[t]
\small

\SetAlgoLined
\KwIn{Training episode sampled $\{\emph{Q} , \emph{S}\}$ from training set $D$ }
 initialize E, $\phi$ (using pre-trained model)\;
 \While{not converged}{
  Sample $N_c$-way $K$-shot episode (\emph{S}, \emph{Q}) from base classes
  
  Encode instance $\phi(\boldsymbol{x}_i)$ and $\phi(\boldsymbol{x}_j)$, for $\boldsymbol{x}_i \in \emph{S}$ and $\boldsymbol{x}_j \in \emph{Q}$
  
  Calculate class summarization as $\boldsymbol{p}_c =\text{SetFunction}(\{\phi(\boldsymbol{x}_i)\}_{i\in \emph{S}_c}$ 
  
  Compute the score: $z_n^c=\text{cos}(\boldsymbol{p}_c, \boldsymbol{e}_n)$ 
  
  Generate the classifiers with meta-components $\{\boldsymbol{e}_n\}$: $\boldsymbol{w}_c = \boldsymbol{E}^\top \boldsymbol{\zeta}_c$ 

    Predict $\hat{y}$ and evaluate the loss on the query $\ell_\emph{Q}$
    
    Update $\phi$ and $\boldsymbol{E}$ by gradient descent steps }
\caption{MCL}
\label{algo}
\end{algorithm}

  
  

\end{appendices}
\end{document}